%% file: iclr2026_conference.tex
\documentclass{article} 
\usepackage{iclr2026_conference,times}

\input{math_commands.tex}

\usepackage{hyperref}
\usepackage{url}

\usepackage{graphicx} 
\PassOptionsToPackage{numbers, compress}{natbib}
\usepackage{colortbl}
\usepackage{graphicx}
\usepackage{wrapfig}
\usepackage{hyperref}
\hypersetup{colorlinks,linkcolor={red},citecolor={blue}, urlcolor=orange}  
\definecolor{COLOR_MEAN}{HTML}{f0f0f0}
\definecolor{LIGHT_BLUE}{HTML}{e6f1fe}
\definecolor{LIGHT_RED}{HTML}{fceeee}
\definecolor{LIGHT_YELLOW}{HTML}{f1f58a}
\definecolor{LIGHT_GREEN}{HTML}{eaffea}
\definecolor{LIGHT_BROWN}{HTML}{f5e6d3}
\usepackage{multirow} 
\usepackage{makecell} 
\usepackage{caption}
\usepackage{cleveref}
\usepackage{tcolorbox}
\usepackage[subrefformat=parens]{subcaption}
\tcbuselibrary{breakable}
\usepackage{fontawesome}
\usepackage{url}
\usepackage{stfloats} 
\usepackage{booktabs}
\usepackage{xspace}
\newcommand{\ie}{\textit{i.e.},\xspace}

\title{Task-CoEvolve: Efficient Harness Optimization via Adaptive Validation Task Selection}

\author{
Atsuyuki Miyai\hspace{0.8em}
Kiyoharu Aizawa\thanks{Co-supervised this work.}\hspace{0.8em}
Toshihiko Yamasaki\footnotemark[1] \\[1ex]
The University of Tokyo \\
\texttt{miyai@cvm.t.u-tokyo.ac.jp}
}
\iclrfinalcopy 
\begin{document}

\maketitle

\begin{figure*}[h]
\vspace{-25pt}
\centering
    \includegraphics[width=0.95\linewidth]{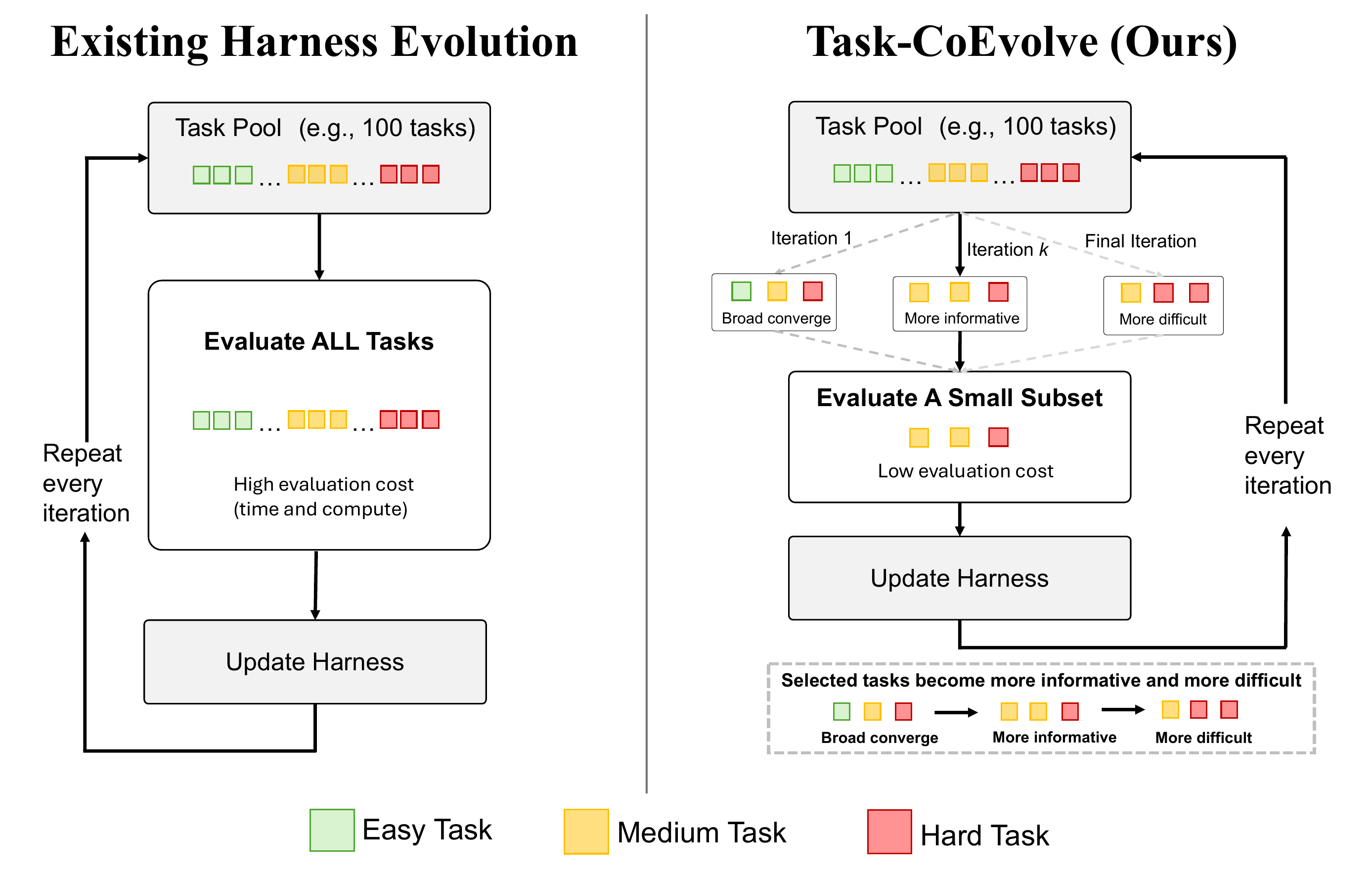}\\
    \vspace{-5pt}
   \caption{\textbf{Comparison between existing harness optimization and our Task-CoEvolve.}}
    \label{fig:fig_teaser}
\end{figure*}

\begin{abstract}
We present a novel approach to efficient LLM harness optimization through adaptive validation task selection. Harness optimization iteratively rewrites the harness code based on validation performance, enabling substantial performance gains without updating the underlying model weights. Existing approaches, however, evaluate a fixed validation set in full at every iteration, incurring substantial evaluation costs even on tasks that become less discriminative as the harness evolves. We propose \textbf{Task-CoEvolve}, which \emph{co-evolves} the validation tasks with the harness by addressing two challenges: selecting informative tasks and estimating full-set performance from partial evaluations. Task-CoEvolve builds on the observation that tasks on which candidate harnesses disagree are more informative for distinguishing among them than tasks that are consistently solved or failed. It uses variance-weighted sampling based on past outcomes to focus evaluation on tasks near the capability frontier, with the sampling distribution adapting as the harness evolves. It then estimates full-set scores from the sampled tasks by accounting for their sampling probabilities, enabling consistent comparisons across iterations despite evaluating different subsets. Experiments on online text classification and Terminal-Bench 2.1 show that Task-CoEvolve consistently outperforms subset-based baselines and matches the final performance of full-set search while reducing the number of evaluations during optimization by 80\%. Code will be released at \url{https://github.com/Agent4Science-UTokyo/Task-CoEvolve}.
\end{abstract}

\section{Introduction}

When deploying LLM agents to real-world tasks, the design of the \textit{harness}, the code that determines what to store, what to retrieve, and what to present to the model, is critical. Indeed, simply changing the harness around a fixed LLM has been shown to yield up to a $6\times$ difference in performance on the same benchmark~\citep{Tian2026SWEBenchMC}, suggesting that the harness can matter as much as the underlying model itself. Traditionally, harnesses have been designed by hand~\citep{zhang2025ace, ye2026meta, merrill2026terminal, terminuskira2026}, but their vast design space makes this process costly and reliant on extensive trial and error. Recently, \emph{automated harness optimization} has begun to attract considerable attention, where a meta-level agent iteratively rewrites the harness code, evaluates the resulting candidates on a benchmark, and retains changes that improve performance~\citep{lee2026meta, lin2026agentic, zhang2026harnessing, wang2026rethinking, guo2026drevo,  weng2026harness}.

Existing work on automated harness optimization, however, has largely adopted a naive evaluation strategy: at every iteration, each candidate harness is evaluated on the entire \emph{fixed} set of validation tasks~\citep{lee2026meta, lin2026agentic, zhang2026harnessing, wang2026rethinking}. While simple to implement and straightforward for comparing candidates, this strategy has two key limitations. First, it is expensive. Every iteration requires inference over the full validation set, and when individual tasks are costly to execute, such as long-horizon terminal tasks that occupy a sandbox environment for tens of minutes~\citep{merrill2026terminal}, evaluation can dominate the cost of the entire optimization loop. Second, it is \emph{static}. As the harness evolves, the tasks that meaningfully discriminate among candidates also change. Tasks that every candidate can already solve, or that no candidate can yet solve, continue to consume the evaluation budget while providing little signal for optimization. Overcoming these limitations requires balancing evaluation cost against the informativeness of the resulting optimization signal and adapting the validation tasks themselves as the harness evolves.

In this paper, we study efficient harness optimization using only a subset of validation tasks at each iteration. We propose \textbf{Task-CoEvolve} that \emph{co-evolves} the validation task set alongside the harness (\Cref{fig:fig_teaser}). Task-CoEvolve addresses two challenges: (i) selecting tasks that best discriminate among candidate harnesses and (ii) making evaluations comparable across iterations despite using different subsets.
For (i), we observe that tasks where candidates have different outcomes are more informative, while tasks that are always solved or failed provide little information for ranking candidates. We measure this using the Bernoulli variance of each task's historical success rate, which becomes large when success and failure are balanced. As the harness evolves, the sampling distribution also changes to focus on informative tasks for the current harness. For (ii), we use task inclusion probabilities to estimate full-set scores from each sampled subset, providing a common evaluation criterion across iterations. Together, these components concentrate evaluation on informative tasks while enabling fair comparison and final selection without evaluating the full validation set at every iteration.

Following prior work~\citep{lee2026meta}, we evaluate Task-CoEvolve on online text classification~\citep{fei2024lawbench, gretel2023symptom, schneider2016s} and Terminal-Bench 2.1~\citep{merrill2026terminal}, a benchmark for long-horizon terminal agents. On text classification, Task-CoEvolve approaches full-set search even with an extreme evaluation budget of only 7\%. With a 20\% evaluation budget, Task-CoEvolve even outperforms full-set search. On Terminal-Bench 2.1, Task-CoEvolve matches the performance of full-set search using only 20\% of the evaluations, while reducing the overall search cost by 67-80\%.
Our contributions are summarized as follows:
\begin{itemize}
\item \textbf{Problem Setting.} We introduce the problem of optimizing \emph{which tasks} are used to evaluate candidate harnesses. This direction is orthogonal to prior efficiency approaches that reduce the number of candidates, instead reducing the evaluation cost per candidate.

\item \textbf{Task-CoEvolve.} We propose Task-CoEvolve, which combines adaptive task selection based on discriminative power with full-set score estimation from partial evaluations (\Cref{fig:fig_teaser}). The former concentrates the evaluation budget near the capability frontier, while the latter provides a common evaluation criterion across different subsets.

\item \textbf{Empirical Findings.} On text classification, Task-CoEvolve approaches full-set search with only 7\% of the evaluation budget and surpasses it with 20\% (\Cref{table:tc}). On Terminal-Bench 2.1, it achieves comparable performance while reducing search costs by 67--80\%.

\end{itemize}

\section{Related Work}
\label{sec:related_work}

\textbf{Automatic Optimization of Harnesses.}
Research on recursive self-improvement has traditionally focused on improving the model itself by updating its weights~\citep{huang-etal-2023-large, pmlr-v235-yuan24d, NEURIPS2025_6b41e04c, NEURIPS2022_639a9a17}. More recently, attention has shifted toward improving the harness while keeping the underlying model fixed~\citep{lee2026meta, lin2026agentic, nie2026tthe, zhang2026harnessing}. These approaches progressively improve the harness through an iterative loop of proposing harness modifications, evaluating the resulting candidates, and adopting promising changes. A representative example is Meta-Harness~\citep{lee2026meta}, which evaluates multiple harness candidates on a given task set and leverages their execution traces and performance histories to iteratively improve the harness, ultimately aiming to maximize performance over the target task distribution. A common limitation of these approaches~\citep{lee2026meta, lin2026agentic, zhang2026harnessing}, however, is that they repeatedly evaluate candidate harnesses on the full validation task set at every iteration. As a result, evaluation can incur substantial computational and time costs, particularly when individual tasks are expensive to execute.

\textbf{Efficient Harness Optimization.}
Recent work has explored improving the efficiency of harness and program optimization from several directions. DemoEvolve~\citep{che2026demoevolve} incorporates human demonstrations to provide more informative feedback in sparse-reward settings, while ShinkaEvolve~\citep{lange2025shinkaevolve} and TurboEvolve~\citep{yang2026turboevolve} improve the sample efficiency of LLM-driven evolutionary search through more efficient candidate generation and selection strategies. HarnessCompass~\citep{zhang2026harnesscompass} further improves harness evolution through constrained, feedback-guided, and component-wise optimization. These approaches primarily improve efficiency on the search side, such as by generating, selecting, or refining candidate harnesses more effectively. Our work addresses an orthogonal source of cost: the number of tasks used to evaluate each candidate. Task-CoEvolve complements these approaches by reducing the per-candidate evaluation cost.

\textbf{Curriculum Learning and Adaptive Task Selection.}
Curriculum learning and adaptive task selection aim to improve learning efficiency and performance by dynamically selecting training tasks according to the model's current capabilities~\citep{bengio_curriculum, ruvolo2013active, soviany2022curriculum}. Broadly, these methods optimize \emph{what to learn from}. In harness optimization, by contrast, the key question is \emph{what to evaluate on}. While training tasks provide supervision for updating model parameters, evaluation tasks in harness optimization determine which candidate harnesses are preferred and thereby guide the direction of optimization.

\textbf{Sample-Efficient Model Evaluation.}
Prior work has reduced evaluation costs by estimating model performance from selected test examples. Active testing selectively evaluates informative test examples with importance-weighted performance estimation~\citep{kossen21a2021active}. tinyBenchmarks~\citep{polo2024tiny} and AcTracer~\citep{huang2026actracer} further improve sample-efficient LLM evaluation through compact benchmark subsets and model-informed sampling, respectively.
These approaches primarily focus on efficient performance estimation of a fixed model. In contrast, Task-CoEvolve selects tasks to discriminate among evolving harness candidates, while accounting for non-uniform sampling when estimating full-set performance.

\begin{figure*}[t]
\vspace{-25pt}
\centering
    \includegraphics[width=0.95\linewidth]{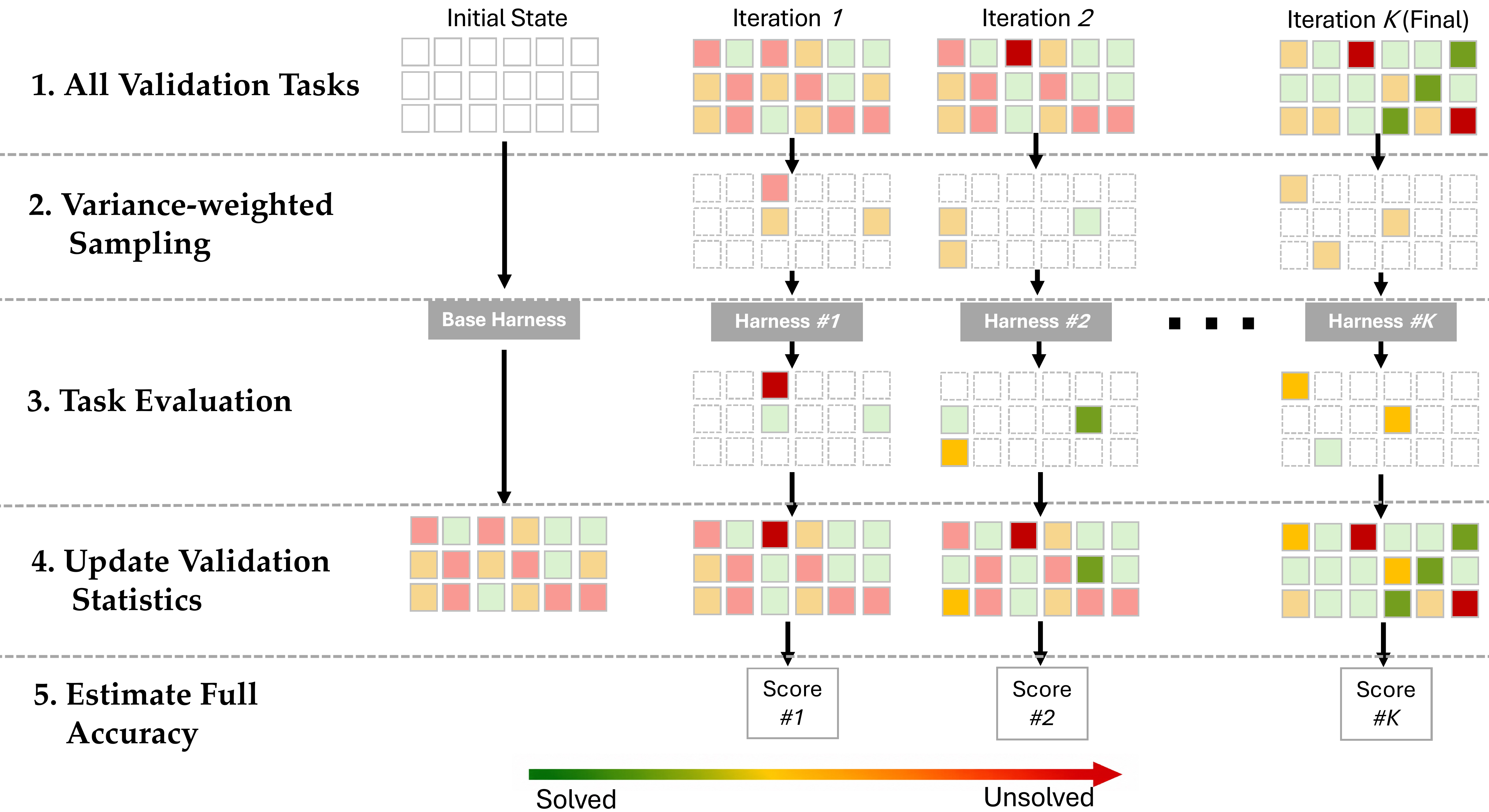}\\
    \vspace{-5pt}
\caption{\textbf{Overview of the Task-CoEvolve procedure.} Task-CoEvolve first selects a subset of tasks from the full validation set using variance-weighted sampling and evaluates the candidate harness on the selected tasks. The resulting outcomes are used to update the validation-task statistics. Finally, Task-CoEvolve estimates the full-set accuracy from the sampled evaluations.}
    \label{fig:overview}
\end{figure*}

\section{Method}
\label{sec:method}

\subsection{Problem Statement}

We consider a fixed LLM and a \textit{harness} $h$, \ie the control code surrounding the model. Let $\mathcal{T}=\{1,\dots,N\}$ denote a set of validation tasks, and let $x_t(h)\in[0,1]$ denote the average success rate of harness $h$ on task $t\in\mathcal{T}$ over $r$ trials. We define the true performance of a harness $h$ as its full-set score
\begin{equation}
\begin{aligned}
S(h) = \frac{1}{N} \sum_{t \in \mathcal{T}} x_t(h).
\end{aligned}
\label{eq:full_set_score}
\end{equation}

In harness optimization, a meta-level agent proposes a candidate harness $h_k$ at each iteration $k=1,\dots,K$ based on previous evaluation results, and feeds the new results back into subsequent iterations. The goal is to select, among all candidates generated during optimization, the harness with the highest $S(h)$. In this paper, we consider achieving this goal under a limited evaluation budget, measured by the total number of task executions.

Importantly, our task selection and full-set estimation procedures depend only on observed task outcomes, \ie the history of successes and failures, and make no assumptions about either the task content or the harness code.

\subsection{Review of Meta-Harness}
\label{subsec:review_meta_harness}

Meta-Harness~\citep{lee2026meta} is a representative framework for automated harness optimization. At each iteration, the meta-agent receives previous harness candidates and their evaluation results as context and generates the code for a new harness $h_k$. The candidate $h_k$ is then evaluated on the \emph{entire} validation task set $\mathcal{T}$, and its average score, $\frac{1}{N}\sum_{t \in \mathcal{T}} x_t(h_k)$, is recorded. After $K$ iterations, the candidate with the highest recorded score is selected as the final harness. This procedure requires a total of $K \times N \times r$ task executions, incurring an evaluation cost proportional to $N$ at every iteration.
\subsection{Proposed Approach: Task-CoEvolve}
\label{subsec:proposed_approach}

A natural way to reduce evaluation cost is to evaluate each candidate on only a subset $\mathcal{S}_k \subset \mathcal{T}$ of size $m = \lceil \rho N \rceil$, which reduces the task executions from $K \times N \times r$ to $K \times m \times r$. This raises two challenges. Reusing a fixed subset invites overfitting, as the meta-agent keeps favoring modifications that work on the same tasks. Resampling it every iteration avoids this, but the raw subset mean then depends on the difficulty of the sampled tasks, so scores from different iterations are no longer comparable.

\textbf{Overall Concept.}
Task-CoEvolve addresses these two challenges with two components. First, \emph{variance-weighted task selection} selects a new validation subset at each iteration based on past evaluation outcomes, focusing the evaluation budget on informative tasks. Second, \emph{full-set score estimation} estimates the full-set performance from the sampled tasks while accounting for their sampling probabilities~\citep{Horvitz01121952}. This allows us to compare candidates evaluated on different subsets using the same scale. \Cref{fig:overview} shows the overall procedure.

\textbf{Phase 0: Initialization.}
Before the search starts, we evaluate two starting harnesses on the full task set $\mathcal{T}$. These runs are not an extra cost of our method: harness optimization needs them in any case, because the meta-agent writes its first candidate on top of them. We reuse their outcomes as the initial history, so that every task already has a success rate $\bar{p}_t$ when the first subset is drawn.

\textbf{Phase 1: Variance-Weighted Task Selection.}
We use a simple criterion to measure how informative each task is: the Bernoulli variance of its past outcomes. This variance becomes large when previous harnesses have different outcomes on the task. At iteration $k$, let $\bar{p}_t$ and $n_t$ denote the mean and the number of outcomes observed for task $t$ in previous iterations. We define the sampling weight for task $t$ as follows:

\begin{equation}
\begin{aligned}
w_t = \max\bigl( \bar{p}_t (1 - \bar{p}_t),\; \ell_t \bigr) + \frac{\lambda}{\sqrt{n_t}}.
\end{aligned}
\label{eq:vws_weight}
\end{equation}
The first term represents the Bernoulli variance, which is largest at $\bar{p}_t=0.5$ and becomes zero when the task is always solved or always failed. The floor $\ell_t$ equals a small positive constant $\ell$ for tasks that have never been solved and $0$ otherwise, because a task nobody has solved may still become solvable as the harness improves. The second term, $\lambda/\sqrt{n_t}$, gives a larger weight to tasks with fewer observations, so that they are not excluded based on only a few early outcomes.

\textbf{Phase 2: Sampling-Aware Full-Set Estimation.}
To enable fair comparisons across iterations, we estimate full-set performance from the sampled tasks. The key idea is to account for each task's inclusion probability $\pi_t = \Pr[t \in \mathcal{S}_k]$ under the sampling design~\citep{Horvitz01121952}, which we estimate via Monte Carlo simulation. However, we find that the appropriate form of full-set estimation depends on the structure of the task pool in each benchmark (refer to \Cref{app:estimator_swap} for detailed experiments). Therefore, guided by the Phase-0 evaluations and the type of task set, we use one of the following two estimators for each benchmark.

\textbf{\textit{H\'ajek estimation when success rates sit near $0$ or $1$.}}
Weighting each sampled outcome by $1/\pi_t$ gives the H\'ajek estimator~\citep{hajek1964asymptotic}
\begin{equation}
\begin{aligned}
\hat{S}(h) = \frac{\sum_{t \in \mathcal{S}_k} x_t(h) / \pi_t}{\sum_{t \in \mathcal{S}_k} 1 / \pi_t}.
\end{aligned}
\label{eq:full_estimate}
\end{equation}
This form suits a task set that (i) splits into several pools, each sampled and estimated separately and then averaged, and (ii) has a mean success rate close to $0$ or $1$ within every pool, since a rarely sampled task then carries an outcome close to its pool mean. 

\textbf{\textit{Anchored difference estimation when they sit near the middle.}}
Otherwise a consistently solved task with a tiny $\pi_t$ is occasionally sampled and its term $x_t(h)/\pi_t$ dominates \Cref{eq:full_estimate}. We therefore weight not the outcome but its deviation from an \emph{anchor} $\bar{p}_t$, the historical success rate of task $t$:
\begin{equation}
\begin{aligned}
\hat{S}(h) = \frac{1}{N}\sum_{t \in \mathcal{T}} \bar{p}_t \;+\; \frac{1}{N}\sum_{t \in \mathcal{S}_k} \frac{x_t(h) - \bar{p}_t}{\pi_t}.
\end{aligned}
\label{eq:difference}
\end{equation}
The anchor is fixed before sampling, so it cancels in expectation.

\textbf{Phase 3: Final Selection.}
After optimization, we select the candidate with the highest estimated full-set score $\hat{S}(h)$ as the final harness, breaking exact ties in favor of the earliest iteration.
We provide additional experiments on the tie-breaking rule in \Cref{app:tiebreak}.

\section{Experiments}
\label{sec:experiments}

Following \citet{lee2026meta}, we evaluate in two settings. We use online text classification for rigorous verification of Task-CoEvolve, precisely quantifying its gains through comprehensive experiments. We use Terminal-Bench 2.1~\citep{merrill2026terminal} to test whether our findings generalize to a realistic, compute-intensive setting.

\subsection{Setup on Online Text Classification}
\textbf{Benchmark and Implementation Details.}
We follow the online text classification setup of \citet{zhang2025ace,ye2026meta,lee2026meta}: the LLM receives labeled examples one at a time, updates its memory (harness), and is evaluated on a held-out test set. Following \citet{lee2026meta}, we use \texttt{GPT-OSS-120B}~\citep{agarwal2025gpt} with temperature $0$ as the classifier LLM, and automatically optimize its harness. We use three datasets spanning different domains and levels of difficulty:
\textbf{LawBench} (Law)~\citep{fei2024lawbench}, which predicts criminal charges from case descriptions (215 classes);
\textbf{Symptom2Disease} (S2D)~\citep{gretel2023symptom}, which predicts diseases from symptom descriptions (22 classes); and
\textbf{USPTO-50k}~\citep{schneider2016s}, which predicts precursor reactants from product molecules (180 classes).
The train/validation/test splits are 200/50/100 for Law, 200/50/212 for S2D, and 50/30/100 for USPTO, yielding a total of $N=130$ validation examples and 412 test examples.

Following Meta-Harness, we use Claude Opus 4.6~\citep{anthropic2026claudeopus46} as the meta-agent and run 20 evolution iterations, generating three candidate harnesses per iteration for a total of 60 candidates. 
To study the effect of evaluation budget, we run each method with validation-set sampling rates $\rho \in {7\%, 20\%}$. For every setting, we perform three runs and report the held-out test accuracy of the selected final harness, averaged across the three datasets.

\textbf{Baselines and Comparison Methods.}
We use Meta-Harness~\citep{lee2026meta} as the basic search framework. Meta-Harness is a representative framework for automatic harness optimization (\Cref{subsec:review_meta_harness}). Its simple structure and high extensibility make it suitable for comparing different evaluation designs. Following the official implementation of Meta-Harness, all evolutionary protocols start from two initial harnesses: zero-shot (direct prompting without memory) and few-shot (all) (putting all training examples into the context).

We consider three evolutionary methods:
\textbf{(1) Meta-Harness (Full Search)} is the original protocol~\citep{lee2026meta}, which evaluates the full validation set ($\rho=100\%$) at every iteration;
\textbf{(2) Naive} samples a fixed subset at the beginning of search and reuses the same subset for all iterations. The candidate with the highest raw subset score is selected.
\textbf{(3) Random-Resample} resamples a subset $\mathcal{S}_k$ randomly from $\mathcal{T}$ at every iteration, but uses the subset score for final selection. This baseline helps separate the effect of simply changing the evaluation subset to reduce overfitting from the effects of discriminative selection and sampling-aware estimation in our method. Scores here are estimated per dataset, and the rule of \Cref{subsec:proposed_approach} assigns the H\'ajek estimator (\Cref{eq:full_estimate}) to this setting (\Cref{table:tc_phase0}).

\subsection{Results on Online Text Classification}
\label{subsec:result_text-classification}
We show the results in \Cref{table:tc}. The main findings are as follows.
\input{tables/tc}

\textbf{Task-CoEvolve achieves the highest accuracy at both budgets.}
Task-CoEvolve achieves 49.3\% at $\rho{=}20\%$ and 47.6\% at $\rho{=}7\%$, outperforming Naive by 2.1 and 2.4 points, respectively. 
With $\rho{=}7\%$, Task-CoEvolve improves the few-shot accuracy from 41.6\% to 47.6\% while using 16 times fewer samples than full-set search.

\textbf{Task-CoEvolve outperforms Meta-Harness at $\rho=20\%$.}
Surprisingly, Task-CoEvolve outperforms Meta-Harness by about 1\% even when using only 20\% of the validation set. One possible reason is that Meta-Harness overfits to the validation set during search, leading to lower performance on the test set. This result suggests that changing the evaluation samples across iterations can reduce the risk of overfitting.

\subsection{Setup on Terminal-Bench 2.1}
\label{subsec:setup_terminal}
\textbf{Benchmark and Implementation Details.}
Terminal-Bench 2.1~\citep{merrill2026terminal}, a minor evaluation update to Terminal-Bench-2, evaluates LLM agents on 89 challenging tasks that require long-horizon, fully autonomous execution under complex dependencies and substantial domain knowledge. Following \citet{lee2026meta}, we use the same 89 tasks for both search and final evaluation. This is because the benchmark is small and expensive enough that introducing a separate split would substantially weaken the search signal~\citep{lee2026meta}.

We use GPT-5.6 Luna~\citep{openai2026gpt56} and Qwen3.6-35B-A3B~\citep{qwen36_35b_a3b} to balance performance and inference cost, as running large-scale evaluations on Terminal-Bench 2.1 is particularly expensive. We set the number of runs per task to $r=1$ (one rollout per task per iteration) and run 10 evolution iterations, generating one candidate harness per iteration and 10 candidates in total.
A small fraction of sandboxed executions can fail randomly due to infrastructure issues unrelated to the harness, introducing noise into the evaluation. We therefore retry such failures at most twice based on a fixed list of errors, using the same rule for all experiments.

\textbf{Baselines and Comparison Methods.}
Following \citet{lee2026meta}, we initialize the search from two strong open baselines, Terminus 2~\citep{merrill2026terminal} and Terminus-KIRA~\citep{terminuskira2026}. We use the same evolutionary comparison methods as in the online text classification experiments. The 89 tasks form a single pool in which most tasks are either always or never solved by the two starting harnesses (\Cref{table:difficulty}), while their mean success rate on the same 89 tasks is near $0.5$, so a rarely sampled task carries an outcome far from the pool mean. Following the rule in \Cref{subsec:proposed_approach}, Task-CoEvolve therefore uses the anchored difference estimator (\Cref{eq:difference}) here.

\subsection{Results on Terminal-Bench 2.1}
\label{subsec:result_terminal}
\input{tables/tb}

We show the results on Terminal-Bench 2.1 in \Cref{table:tb2_results}.
The main findings are as follows.

\textbf{Task-CoEvolve outperforms the comparison methods and is comparable to Full Search.}
As shown in \Cref{table:tb2_results}, Task-CoEvolve outperforms both Naive and Random-Resample with GPT-5.6 Luna and Qwen3.6-35B-A3B. Compared with Meta-Harness (Full Search), Task-CoEvolve is only about 1\% lower in both settings, which corresponds to just one task out of 89. Following prior work, Terminal-Bench uses the validation accuracy itself as the final evaluation result. Therefore, Meta-Harness (Full Search) has an inherent advantage because it can evaluate all tasks at every iteration. Considering this advantage, the small performance gap suggests that Task-CoEvolve achieves performance comparable to Full Search.

\subsection{Search Cost and Time on Terminal-Bench 2.1}
\label{subsec:tb2-cost}
\input{tables/tb_cost}
Terminal-Bench 2.1 requires tens of minutes of sandbox execution per task, making the search process itself a practical bottleneck. We therefore show the actual search cost reduction achieved by our method in \Cref{table:tb2_cost}.

\textbf{Task-CoEvolve reduces the search cost by 67--80\%.}
For GPT-5.6 Luna, Full Search consumes 2{,}888M input tokens (22.2 hours), while Task-CoEvolve requires only 579M tokens (11.5 hours). For Qwen3.6, the input token usage decreases from 741M to 246M, and the search time decreases from 38.0 to 20.5 hours. Despite these substantial reductions, the final performance is only 1.1 points lower than Full Search (\Cref{table:tb2_results}), showing that Task-CoEvolve achieves comparable search results at only one-third to one-fifth of the token cost.

\textbf{The same 20\% evaluation budget can result in very different costs.}
Although all 20\% protocols evaluate the same number of tasks, their token consumption differs substantially. The average input tokens per trial are 0.7M for Random-Resample and 1.2M for Naive, while Task-CoEvolve uses 3.2M, comparable to Full Search (3.2M). This is because variance-weighted selection concentrates the evaluation budget on long-running, multi-turn tasks where candidate harnesses show different success and failure outcomes. Therefore, the large cost reduction of Random-Resample (96\%) should not necessarily be interpreted as higher efficiency; rather, uniform sampling also spends its budget on easier tasks that terminate quickly. Indeed, under the same 20\% evaluation budget, Random-Resample achieves a final performance 3.3 points lower than Task-CoEvolve.

\textbf{Search time is reduced by about half.}
The reduction in search time ($1.9\times$) is smaller than the reduction in token usage ($5.0\times$) for two reasons. First, trials are always executed with 10-way parallelism, so the search time is mainly determined by the longest-running tasks rather than the number of evaluated tasks. Second, candidate proposal by the meta-agent takes about 2.4--3.3 hours regardless of the evaluation budget. This is a fixed cost and therefore becomes more significant as the evaluation cost decreases. Even with this fixed cost, Task-CoEvolve reduces the total search time by about half: from 22.2 to 11.5 hours for GPT-5.6 Luna and from 38.0 to 20.5 hours for Qwen3.6, while maintaining performance close to Full Search.

\section{Analysis}
\subsection{Ablation on Each Component}
\label{subsec:ablation}
\input{tables/ablation_contribution}
We show the contributions of the three components of Task-CoEvolve: subset resampling, design-based estimation with $\hat{S}$-max selection, and variance-weighted adaptive selection in \Cref{table:ablation}, at $\rho{=}20\%$.

\textbf{Random-Resample gives the largest single improvement, but scores are still not comparable.}
Resampling the subset uniformly at random at every iteration improves the average performance from $47.2$ to $48.2$, which is the largest gain among the three components. This suggests that repeatedly using the same subset is a major reason for the poor performance of Naive. However, the raw subset scores are still affected by the difficulty of each sampled subset, making comparisons across iterations unreliable.

\textbf{Estimation and variance-weighted selection further improve performance.}
Adding full-set estimation and $\hat{S}$-max selection (Est.) improves the average performance from $48.2$ to $48.8$, showing the benefit of comparing candidates on a common scale. Adding variance-weighted selection (VWS) further improves the performance to $49.3$, achieving the best result.

\subsection{Analysis of Sample Discriminability}
\label{subsec:difficulty}
\Cref{table:difficulty} divides samples into four groups based on historical accuracy $\bar{p}$: never solved ($\bar{p}=0$), sometimes solved ($0<\bar{p}<1/3$), mixed ($1/3 \leq \bar{p} \leq 2/3$), and mostly solved ($\bar{p}>2/3$).
\input{tables/difficulty}

\textbf{Most of the evaluation budget is spent on samples that cannot distinguish between candidates.}
In both settings, the two extremes ($\bar{p}=0$ and $\bar{p}>2/3$), which provide little information for ranking candidates, consistently account for more than 70\% of the sample pool. In contrast, samples where candidate outcomes are most divided ($1/3 \leq \bar{p} \leq 2/3$) remain a relatively small fraction of the task pool throughout the search. Uniform subset sampling naturally preserves this imbalance.

\textbf{Sample difficulty changes as the harness evolves.}
The distribution is not static. In text classification, the number of samples solved by almost all candidates increases from $34$ to $58$, meaning that previously useful samples become too easy to distinguish between candidates. In Terminal-Bench 2.1, the number of tasks that no candidate can solve decreases from $32$ to $21$, as previously unsolved tasks become solvable. These changes support our motivation for adapting the evaluation samples throughout the search.

\subsection{Accuracy of the Estimated Scores}
To evaluate the accuracy of $\hat{S}$, we re-evaluate all 60 candidates generated during a search run on the full validation set (130 samples) and compare their estimated scores $\hat{S}$ with the true scores (\Cref{table:est_accuracy}).

We find that the role of $\hat{S}$ differs substantially depending on the evaluation budget.

\input{tables/estimate_score}
\textbf{At $\rho=20\%$, $\hat{S}$ provides a useful ranking signal.}
The rank correlation between $\hat{S}$ and the true score is 0.62. The candidate with the highest $\hat{S}$ turns out to be the 12th best of the 60, with a true score of 51.3\% against 54.7\% for the best candidate. So $\hat{S}$ does not find the very best candidate, but it keeps the choice within the top fifth of the pool.

\textbf{At $\rho=7\%$, ranking is difficult, but $\hat{S}$ still prevents poor candidate selection.}
The rank correlation drops to 0.13, since two or three samples per dataset carry little information about a candidate; the limitation comes from the sample size rather than from the estimator. The selected candidate still ranks 10th of 60, at 46.4\% against 47.8\% for the true best.

\section{Conclusion, Limitations and Future Work}

We introduce a new problem setting for automated harness optimization, where we optimize not only the harness itself but also which tasks are used to evaluate candidate harnesses. We propose Task-CoEvolve, which combines adaptive task selection based on discriminability with sampling-aware estimation of full-set performance. On text classification, Task-CoEvolve achieves performance close to Full Search using only a 7\% evaluation budget and even outperforms it at 20\%. On Terminal-Bench 2.1, it reduces the search cost by 67--80\% while achieving performance similar to Full Search.

A limitation is that Task-CoEvolve fixes how many tasks each candidate is evaluated on before seeing any of its results. It therefore cannot stop early on a candidate that is already clearly worse, nor evaluate more tasks when two candidates are hard to tell apart. Deciding this number during evaluation is a natural next step and we leave it to future work.

\section*{Acknowledgments}
We thank Kenta Watanabe and Zhenyu He for their helpful discussions and feedback on this work.

\bibliography{iclr2026_conference}
\bibliographystyle{iclr2026_conference}
\clearpage
\appendix
\newcommand\beginsupplement{%
        \setcounter{table}{0}
        \renewcommand{\thetable}{\Alph{table}}%
        \setcounter{figure}{0}
        \renewcommand{\thefigure}{\Alph{figure}}%
     }
\beginsupplement
\section*{Appendix}

\section{Effect of the Estimator Choice}
\label{app:estimator_swap}
In this section, we swap the full-set estimator used for each benchmark while keeping all other components fixed: the task selection method, evaluation budget, $\hat{S}$-max selection rule, and meta-agent. Thus, the only difference is how the sampled subset is used to estimate the full-set score $\hat{S}$.

\subsection{Difference Estimation on Online Text Classification}
\input{tables/estimator_swap}
\Cref{table:estimator_swap} shows that replacing the estimator with difference estimation degrades performance on online text classification. Mean test accuracy drops by $3.3$ points at $\rho{=}20\%$ and $3.6$ points at $\rho{=}7\%$.
In text classification, each dataset contains only two to ten sampled examples per iteration. Subtracting the historical anchor therefore reduces the differences between candidates, which makes candidate comparison less reliable.

\subsection{H\'ajek Estimation on Terminal-Bench 2.1}
On Terminal-Bench 2.1,replacing the difference estimator with the H\'ajek estimator leads to unstable estimation. In a Qwen3.6 run with $\rho{=}10\%$, each candidate is evaluated on three tasks using \Cref{eq:full_estimate}. At iteration $4$, the candidate solved only one of the three tasks, with a raw score of $33.3\%$, but its estimated score was $\hat{S}=85.9\%$.
\Cref{eq:difference} avoids this problem by using the historical performance as an anchor. For a task that is always solved by the starting harnesses, the anchor is $\bar{p}_t=1$. If a new candidate also solves this task, its residual becomes $x_t(h)-\bar{p}_t=0$, so the task does not affect the estimate.

\section{Sensitivity to the Tie-Break Rule}
\label{app:tiebreak}

Throughout the paper, when multiple candidates have the same selection score, we select the candidate from the earliest iteration, following \Cref{subsec:proposed_approach}. An alternative is to select the latest candidate, since it is generated with more feedback from previous iterations. \Cref{table:tiebreak} compares these two strategies.
\input{tables/tiebreak}
\input{tables/tc_phase0}
\input{tables/parameter}

Task-CoEvolve is not affected by the tie-breaking rule because $\hat{S}$ is continuous, and no candidates share the highest value in our runs. The rule only affects the baselines, which select candidates using raw subset scores. When the subset is small, many candidates can have the same score. For example, at $\rho{=}7\%$, 13--16 out of 60 candidates in a Naive run share the highest score.

The choice of tie-breaking rule does not change the ordering of the methods. In fact, selecting the latest candidate makes the baselines worse: at $\rho{=}7\%$, Naive decreases from $45.2$ to $41.8$, and Random-Resample from $47.0$ to $45.6$. We therefore use the earliest candidate, which is more favorable to the baselines and also follows the implementation of \citet{lee2026meta}, where the first candidate reaching the best score is returned.

\section{A Stronger Model on Terminal-Bench 2.1}
\label{app:stronger_model}
\begin{table}[t]
\centering
\caption{Full-set search with three models. ``Terminus 2'' is the starting harness
and ``Search'' is the best candidate found in ten iterations. The stronger the
model, the less the search has to work with.}
\label{table:stronger_model}
\small
\setlength{\tabcolsep}{6pt}
\begin{tabular}{lccc}
\toprule
Model & Terminus 2 & Search & Gain \\
\midrule
Qwen3.6-35B-A3B & 34.8 & 42.7 & $+7.9$ \\
GPT-5.6 Luna & 52.8 & 62.9 & $+10.1$ \\
DeepSeek-V4-Flash & 70.8 & 70.8 & $+0.0$ \\
\bottomrule
\end{tabular}
\end{table}

The two models in \Cref{table:tb2_results} leave room for harness optimization. We also tested a stronger model, DeepSeek-V4-Flash~\citep{deepseekai2026deepseekv4} under the same full-set search protocol: $89$ tasks, ten iterations, and one candidate per iteration.

\Cref{table:stronger_model} shows the result.
We found that the search in DeepSeek-V4-Flash found no improvement. Terminus 2 already solves $70.8\%$ of the tasks, and the best candidate also reaches $70.8\%$. The other nine score between $64.0\%$ and $69.7\%$. Thus, none of the candidates improves on the starting harness.

This is a property of the benchmark and model, not a limitation of Task-CoEvolve. Stronger models need less scaffolding, leaving little room for harness optimization. Task-CoEvolve can improve how this remaining room is explored, but it cannot create room that does not exist. We therefore use GPT-5.6 Luna and Qwen3.6-35B-A3B in the main experiments, where there is enough headroom for meaningful comparison.

\section{Examples of Discovered Harnesses}
\label{app:examples}

We describe the harness selected by Task-CoEvolve in each setting.

\subsection{Online Text Classification}

The selected harness on this benchmark, at $\rho{=}20\%$, retrieves few-shot
examples with two tokenizers instead of one. It builds two TF-IDF indices over
the stored examples, one on word bigrams and one on character $n$-grams, ranks
the stored examples separately under each, and merges the two rankings by
reciprocal rank fusion:
\begin{equation}
\begin{aligned}
\mathrm{score}(i) = \frac{1}{k + \mathrm{rank}_{\text{bigram}}(i)} + \frac{1}{k + \mathrm{rank}_{\text{char}}(i)}.
\end{aligned}
\label{eq:rrf}
\end{equation}
It then fills the prompt greedily by fused score, dividing the score of an
example by a factor that grows with how many examples of the same label are
already in the prompt, so that a single label cannot occupy the whole context.
\Cref{fig:code_tc} gives the retrieval step in pseudocode.

\begin{figure}[ht]
\small
\begin{verbatim}
  # query q, stored examples E, character budget B, constants k = 60, alpha
  I_word <- tfidf_index(E, tokenizer = word bigrams)
  I_char <- tfidf_index(E, tokenizer = character n-grams)

  r_word <- rank of each e in E by cosine(q, e) under I_word
  r_char <- rank of each e in E by cosine(q, e) under I_char

  for e in E:                                     # reciprocal rank fusion
      s[e] <- 1 / (k + r_word[e]) + 1 / (k + r_char[e])

  P <- []                                         # greedy fill, diversity penalty
  while E is not empty and chars(P) < B:
      e* <- argmax over e in E of  s[e] / (1 + alpha * count(label(e) in P))
      if chars(P) + chars(e*) > B: break
      move e* from E to P
  return P as the few-shot examples of the prompt
\end{verbatim}
\vspace{-6pt}
\caption{Retrieval in the harness selected for online text classification.
Two rankings are fused, then the prompt is filled greedily while down-weighting
labels that are already present.}
\label{fig:code_tc}
\end{figure}

This helps because the three datasets need different kinds of similarity. Word
bigrams work well on legal case descriptions, where a phrase often indicates the
charge. Character $n$-grams work well on molecule strings and on symptom text,
where the useful unit is a substring rather than a word. A single tokenizer must
be a compromise between the two. By fusing the two rankings, the harness can use
the stronger signal for each query, without knowing which dataset the query comes
from. The meta-agent wrote this as its hypothesis before the candidate was
evaluated, and $\hat{S}$ later selected this harness in its run.

\subsection{Terminal-Bench 2.1}

The selected harness on this benchmark, with GPT-5.6 Luna, changes when the agent
stops waiting after it types a command. The stock agent always sleeps for a fixed
time after it sends keystrokes. The evolved harness keeps this wait for keystrokes
that do not run a command, such as a lone \texttt{C-c}. For keystrokes that do run
a command, it checks the terminal and returns as soon as the shell prompt is back.
\Cref{fig:code_tb} gives this in pseudocode.

\begin{figure}[ht]
\small
\begin{verbatim}
  def wait_after(keystrokes, duration):
      if not runs_a_command(keystrokes):          # C-c, a pure wait turn, ...
          send(keystrokes, wait = duration)       # stock behaviour, unchanged
          return

      before <- capture_pane()
      send(keystrokes, wait = min(duration, 0.5))
      prev <- capture_pane()
      delay, elapsed <- 0.5, 0.5

      while elapsed < duration:                   # poll with backoff
          sleep(delay); elapsed <- elapsed + delay
          delay <- min(1.5 * delay, 3.0)
          pane <- capture_pane()
          if pane != before                       # the command has echoed
             and pane == prev                     # the terminal is quiet
             and last_nonblank_line(pane) ends in "#" or "$":
              return                              # prompt is back: stop waiting
          prev <- pane
      return                                      # fell back to the full wait

  def runs_a_command(k):
      return k is a literal Enter key, or k ends in a newline
\end{verbatim}
\vspace{-6pt}
\caption{The wait after a keystroke batch in the harness selected for
Terminal-Bench 2.1. All three conditions must hold before the wait is cut short,
so pagers, REPLs and silent long compiles keep the full duration.}
\label{fig:code_tb}
\end{figure}

This helps because each task has a time budget, and the agent fails if it runs out
of turns. Before proposing the change, the meta-agent measured the waste in the
search log: the fixed sleeps used a quarter to a third of the budget, and about one
fifth of the trials ended in a timeout instead of a wrong answer.

\section{Implementation Details}
\label{app:implementation}

\subsection{Phase-0 Validation Pool for Online Text Classification}
\label{app:tc_phase0}
The estimator rule of \Cref{subsec:proposed_approach} is fixed from the two
starting harnesses evaluated on the full validation set in Phase 0, before any
candidate is generated. \Cref{table:tc_phase0} reports that evaluation. With two
systems, $\bar{p}$ takes only the values $0$, $0.5$ and $1$. Every dataset's mean
$\bar{p}$ lies close to $0$ or $1$, and the samples that receive a tiny inclusion
probability are exactly those at the majority extreme, whose outcomes are close to
that mean. The unanchored estimator is therefore safe here, in contrast to
Terminal-Bench 2.1 (\Cref{table:difficulty}).

\subsection{Hyperparameters of Experiments}
\label{app:vws_hparams}

\Cref{table:vws_hparams} lists the hyperparameters.

\section{The Use of Large Language Models}
\label{app:llm_usage}

We describe how we used large language models (LLMs) in this paper.

\textbf{Research Ideation.} The authors developed the main idea of this paper: making the validation process adaptive so that it evolves as the harness changes. LLMs were not used in developing this idea.

\textbf{Method Design.} We used LLMs to discuss and compare design options for some components, especially when choosing between estimators. The authors made the final decisions, favoring methods with theoretical guarantees, and verified the properties and claims presented in the paper.

\textbf{Implementation.} We used AI coding assistants to help write parts of the experimental code. The authors reviewed and tested all AI-assisted code and verified the experimental runs that produced the results reported in the paper.

\textbf{Writing.} The authors wrote the manuscript with assistance from LLMs, mainly to improve wording and clarity. The authors determined the scientific content and claims, reviewed all LLM-assisted text, and take full responsibility for the final manuscript.

The authors reviewed all AI-assisted work and take responsibility for the final content of this paper.

\end{document}

%% file: math_commands.tex
\usepackage{amsmath,amsfonts,bm}

\def\eqref#1{equation~\ref{#1}}

\def\1{\bm{1}}

\DeclareMathAlphabet{\mathsfit}{\encodingdefault}{\sfdefault}{m}{sl}
\SetMathAlphabet{\mathsfit}{bold}{\encodingdefault}{\sfdefault}{bx}{n}



%% file: tables/tc.tex
\begin{table}[t]
\centering
\caption{\textbf{Online text classification.} Held-out test accuracy of the harness selected by each protocol under evaluation budget $\rho$. ``Val/iter'': validation samples per iteration (USPTO/S2D/Law); ``Evals'': total sample evaluations during search.
\dag: values reported in \citet{lee2026meta}.
Task-CoEvolve nearly matches full-set search at $\rho{=}7\%$ and surpasses it at $\rho{=}20\%$.}
\label{table:tc}
\small
\setlength{\tabcolsep}{4.5pt}
\begin{tabular}{llccc|ccc|c}
\toprule
\textbf{Method} & {\boldmath$\rho$} & {\boldmath$n$} & \textbf{Val/iter} & \textbf{Evals} & \textbf{USPTO} & \textbf{S2D} & \textbf{Law} & \textbf{Avg} \\
\midrule
\multicolumn{9}{c}{\textit{Hand-designed harnesses}} \\
\midrule
Zero-shot & -- & 1 & -- & -- & 13.0 & 66.0 & 9.0 & 29.3 \\
Few-shot & -- & 1 & -- & -- & 15.0 & 84.9 & 25.0 & 41.6 \\
MCE~\citep{ye2026meta}\textsuperscript{\dag} & -- & 1 & -- & -- & 14.0 & 83.0 & 23.0 & 40.0 \\
ACE~\citep{zhang2025ace}\textsuperscript{\dag} & -- & 1 & -- & -- & 16.0 & 77.8 & 29.0 & 40.9 \\
\midrule[\heavyrulewidth]
\multicolumn{9}{c}{\textit{Evolved harnesses (harness search)}} \\
\midrule
Meta-Harness (Full search) & 100\% & 3 & 30/50/50 & 7{,}800 & 17.0\scriptsize$\pm$1.0 & 88.5\scriptsize$\pm$0.7 & 40.3\scriptsize$\pm$3.8 & 48.6\scriptsize$\pm$0.8 \\
\midrule
Naive & 7\% & 3 & 2/3/3 & 480 & 11.0\scriptsize$\pm$7.8 & 85.7\scriptsize$\pm$2.0 & 39.0\scriptsize$\pm$3.6 & 45.2\scriptsize$\pm$3.2 \\
Random-Resample & 7\% & 3 & 2/3/3 & 480 & 18.3\scriptsize$\pm$3.1 & 87.1\scriptsize$\pm$0.3 & 35.7\scriptsize$\pm$9.9 & 47.0\scriptsize$\pm$2.2 \\
Task-CoEvolve (ours) & 7\% & 3 & 2/3/3 & 480 & 16.3\scriptsize$\pm$2.3 & 86.2\scriptsize$\pm$1.8 & 40.3\scriptsize$\pm$0.6 & \textbf{47.6\scriptsize$\pm$0.9} \\
\midrule
Naive & 20\% & 3 & 6/10/10 & 1{,}560 & 14.3\scriptsize$\pm$0.6 & 86.8\scriptsize$\pm$1.7 & 40.3\scriptsize$\pm$0.6 & 47.2\scriptsize$\pm$0.6 \\
Random-Resample & 20\% & 3 & 6/10/10 & 1{,}560 & 16.0\scriptsize$\pm$1.0 & 87.7\scriptsize$\pm$1.2 & 41.0\scriptsize$\pm$3.0 & 48.2\scriptsize$\pm$0.5 \\
Task-CoEvolve (ours) & 20\% & 3 & 6/10/10 & 1{,}560 & 19.3\scriptsize$\pm$1.2 & 86.2\scriptsize$\pm$1.0 & 42.3\scriptsize$\pm$2.5 & \textbf{49.3\scriptsize$\pm$0.8} \\
\bottomrule
\end{tabular}
\end{table}

%% file: tables/tb.tex
\begin{table}[t]
\centering
\caption{\textbf{Terminal-Bench 2.1.} Full-89 task pass rate (\%) of the harness
selected by each protocol. ``$\rho$'':
evaluation budget as a fraction of the 89-task pool per iteration.
Task-CoEvolve at $\rho{=}20\%$ nearly matches full-set search on both models
while using $5\times$ fewer evaluations, and beats both subset baselines.}
\label{table:tb2_results}
\small
\setlength{\tabcolsep}{6pt}
\begin{tabular}{lcccc}
\toprule
\textbf{Method} & {\boldmath$\rho$} & \textbf{GPT-5.6 Luna} & \textbf{Qwen3.6-35B-A3B} & \textbf{Avg} \\
\midrule
\multicolumn{5}{c}{\textit{Starting harnesses (hand-designed)}} \\
\midrule
Terminus-KIRA & -- & 49.4 & 32.6 & 41.0 \\
Terminus 2 & -- & 52.8 & 34.8 & 43.8 \\
\midrule[\heavyrulewidth]
\multicolumn{5}{c}{\textit{Evolved harnesses (harness search)}} \\
\midrule
Meta-Harness (Full search) & 100\% & 62.9 & 42.7 & 52.8 \\
\midrule
Naive & 20\% & 55.1 & 39.3 & 47.2 \\
Random-Resample & 20\% & 59.6 & 37.1 & 48.4 \\
Task-CoEvolve (ours) & 20\% & \textbf{61.8} & \textbf{41.6} & \textbf{51.7} \\
\bottomrule
\end{tabular}
\end{table}

%% file: tables/tb_cost.tex
\begin{table}[t]
\centering
\caption{\textbf{Cost of the search phase on Terminal-Bench 2.1.} ``vs.\ Full'' is the reduction in input tokens relative to
full search. Qwen3.6 was served locally, so no API cost applies. Running time is measured including candidate generation, with 10 trials executed in parallel. \emph{Trials}: 1 or 2 trials fail before counts. This skip does not change the conclusion.}
\label{table:tb2_cost}
\small
\setlength{\tabcolsep}{8pt}
\begin{tabular}{lcccccc}
\toprule
Method & Trials & Input (M) & Output (M) & vs.\ Full & Cost (USD) & Time (h) \\
\midrule
\multicolumn{7}{c}{\textit{GPT-5.6 Luna}} \\
Full search & 890 & 2{,}888 & 17.0 & -- & 117 & 22.2 \\
Naive & 180 & 223 & 2.3 & $-$92\% & 12 & 11.9 \\
Random-Resample & 180 & 124 & 2.1 & $-$96\% & 8 & 9.6 \\
Task-CoEvolve (ours) & 180 & 579 & 5.7 & $-$80\% & 30 & 11.5 \\
\midrule
\multicolumn{7}{c}{\textit{Qwen3.6-35B-A3B (self-hosted)}} \\
Full search & 890 & 741 & 24.6 & -- & -- & 38.0 \\
Naive & 180 & 178 & 6.7 & $-$76\% & -- & 15.1 \\
Random-Resample & 180 & 215 & 4.5 & $-$71\% & -- & 14.4 \\
Task-CoEvolve (ours) & 180 & 246 & 4.8 & $-$67\% & -- & 20.5 \\
\bottomrule
\end{tabular}
\end{table}

%% file: tables/ablation_contribution.tex
\begin{table}[t]
\centering
\caption{Ablation on the contribution of each Task-CoEvolve component.
We set $\rho$ to 20\%. Each component is added on top of the row above.
Mean$\pm$SD of the held-out test accuracy over $n$ runs.}
\label{table:ablation}
\small
\setlength{\tabcolsep}{5pt}
\begin{tabular}{lcccccc}
\toprule
Configuration & Rot. & Est. & VWS & $n$ & Mean \\
\midrule
Naive (fixed subset) &  &  &  & 3 & 47.2\scriptsize$\pm$0.6 \\
+ Random-Resample (uniform, raw val) & \checkmark &  &  & 3 & 48.2\scriptsize$\pm$0.5 \\
+ Full estimation + $\hat{S}$-max & \checkmark & \checkmark &  & 3 & 48.8\scriptsize$\pm$1.4 \\
+ Variance-weighted selection (Task-CoEvolve) & \checkmark & \checkmark & \checkmark & 3 & \textbf{49.3\scriptsize$\pm$0.8} \\
\bottomrule
\end{tabular}
\end{table}

%% file: tables/difficulty.tex
\begin{table}[t]
\centering
\caption{Distribution of task solvability across iterations. We show how the number of validation tasks solved by the harness changes over the optimization iterations. Iteration 0 corresponds to the two starting harnesses.}
\label{table:difficulty}
\small
\setlength{\tabcolsep}{6pt}
\begin{tabular}{lcccc}
\toprule
\textbf{Online text classification} (7\%) & \textbf{iter 0} & \textbf{iter 4} & \textbf{iter 10} & \textbf{iter 20} \\
\midrule
Nobody solves it & 74 & 45 & 45 & 45 \\
A few systems solve it & 0 & 23 & 20 & 14 \\
Roughly half solve it (most discriminative) & 22 & 14 & 10 & 13 \\
Almost everybody solves it & 34 & 48 & 55 & 58 \\
\midrule[\heavyrulewidth]
\textbf{Terminal-Bench 2.1} (GPT-5.6 Luna, 20\%) & \textbf{iter 0} & \textbf{iter 4} & \textbf{iter 7} & \textbf{iter 10} \\
\midrule
Nobody solves it & 32 & 26 & 23 & 21 \\
A few systems solve it & 0 & 3 & 3 & 4 \\
Roughly half solve it (most discriminative) & 23 & 22 & 21 & 21 \\
Almost everybody solves it & 34 & 38 & 42 & 43 \\
\bottomrule
\end{tabular}
\end{table}

%% file: tables/estimate_score.tex
\begin{wraptable}{r}{0.5\textwidth}
\vspace{0pt}
\centering
\caption{\textbf{Accuracy of $\hat{S}$ vs.\ full-val ground truth.}
Spearman: rank correlation between $\hat{S}$ and true scores.
Bottom rows: the $\hat{S}$-max winner's rank and full-val score when all
candidates are re-ranked by true score, next to the true best candidate's
score.}
\label{table:est_accuracy}
\small
\setlength{\tabcolsep}{3pt}
\begin{tabular}{lcc}
\toprule
 & $\rho{=}7\%$ & $\rho{=}20\%$ \\
\midrule
Spearman ($\hat{S}$ vs.\ true) & 0.13 & 0.62 \\
Winner rank by true score & 10 / 60 & 12 / 60 \\
True score: winner / best & 46.4 / 47.8 & 51.3 / 54.7 \\
\bottomrule
\end{tabular}
\vspace{-8pt}
\end{wraptable}

%% file: tables/estimator_swap.tex
\begin{table}[b!]
\centering
\caption{\textbf{Estimator swap on online text classification.}
Both settings use the same search procedure and $\hat{S}$-max selection rule, differing only in the estimator. Using the difference estimator lowers the mean test accuracy.}
\label{table:estimator_swap}
\small
\setlength{\tabcolsep}{6pt}
\begin{tabular}{llcc}
\toprule
$\rho$ & Estimator & Test accuracy \\
\midrule
20\% & H\'ajek (Eq.~\ref{eq:full_estimate}) & \textbf{49.3\scriptsize$\pm$0.8} \\
 & Difference (Eq.~\ref{eq:difference}) & 46.0\scriptsize$\pm$2.6 \\
\midrule
7\% & H\'ajek (Eq.~\ref{eq:full_estimate}) & \textbf{47.6\scriptsize$\pm$0.9} \\
 & Difference (Eq.~\ref{eq:difference}) & 44.0\scriptsize$\pm$3.0 \\
\bottomrule
\end{tabular}
\end{table}

%% file: tables/tiebreak.tex
\begin{table}[t]
\centering
\caption{\textbf{Sensitivity to the tie-break rule.} Held-out test accuracy when
exact ties in the selection score are broken towards the earliest iteration (the rule
used throughout the paper) and towards the latest one. ``Ties'' is the number of
candidates sharing the top score, over the three runs of each cell.}
\label{table:tiebreak}
\small
\setlength{\tabcolsep}{6pt}
\begin{tabular}{llccc}
\toprule
$\rho$ & Method & Ties & Earliest & Latest \\
\midrule
7\% & Naive & 13--16 & 45.2\scriptsize$\pm$3.2 & 41.8\scriptsize$\pm$1.0 \\
 & Random-Resample & 2--5 & 47.0\scriptsize$\pm$2.2 & 45.6\scriptsize$\pm$3.7 \\
 & Task-CoEvolve & 1 & \textbf{47.6\scriptsize$\pm$0.9} & \textbf{47.6\scriptsize$\pm$0.9} \\
\midrule
20\% & Naive & 4--12 & 47.2\scriptsize$\pm$0.6 & 47.1\scriptsize$\pm$1.2 \\
 & Random-Resample & 1--2 & 48.2\scriptsize$\pm$0.5 & 47.9\scriptsize$\pm$1.0 \\
 & Task-CoEvolve & 1 & \textbf{49.3\scriptsize$\pm$0.8} & \textbf{49.3\scriptsize$\pm$0.8} \\
\bottomrule
\end{tabular}
\end{table}

%% file: tables/tc_phase0.tex
\begin{table}[t]
\centering
\caption{\textbf{Phase-0 validation pool (online text classification).} Samples at $\bar{p}{=}0$ or $\bar{p}{=}1$ receive a tiny inclusion probability, and in every dataset the majority of them lies on the side nearer the dataset mean, so drawing one of them does not pull the estimate away from that mean.}
\label{table:tc_phase0}
\small
\setlength{\tabcolsep}{6pt}
\begin{tabular}{lccccc}
\toprule
\textbf{Dataset} & \textbf{Zero-shot} & \textbf{Few-shot} & \textbf{Mean }{\boldmath$\bar{p}$} & {\boldmath$\bar{p}=0$} & {\boldmath$\bar{p}=1$} \\
\midrule
USPTO ($n{=}30$) & 3\% & 13\% & 0.08 & 26 & 1 \\
S2D ($n{=}50$) & 62\% & 80\% & 0.71 & 9 & 30 \\
LawBench ($n{=}50$) & 6\% & 22\% & 0.14 & 39 & 3 \\
\bottomrule
\end{tabular}
\end{table}

%% file: tables/parameter.tex
\begin{table}[t]
\centering
\caption{\textbf{Hyperparameters of variance-weighted task selection.} The same
values are used for online text classification and Terminal-Bench 2.1.}
\label{table:vws_hparams}
\small
\setlength{\tabcolsep}{6pt}
\begin{tabular}{llc}
\toprule
\textbf{Symbol} & \textbf{Description} & \textbf{Value} \\
\midrule
$\ell$ & Weight floor for tasks never solved so far & 0.125 \\
$\lambda$ & Uncertainty bonus coefficient in $\lambda/\sqrt{n_t}$ & 0.025 \\
-- & Monte Carlo repetitions for estimating $\pi_t$ & 4{,}000 \\
-- & History window for computing $\bar{p}_t$ & all past iterations \\
$\rho$ & Sampling rate (evaluation budget) & 7\%, 20\% \\
\bottomrule
\end{tabular}
\end{table}